%% file: main.tex
\documentclass{sig-alternate}
\newcommand{\ignore}[1]{}
\usepackage[pass]{geometry}
\usepackage{fancyhdr}
\usepackage[normalem]{ulem}
\usepackage[hyphens]{url}
\usepackage{graphicx}
\usepackage{amsmath}
\usepackage{colortbl}
\usepackage{changepage}

\usepackage[bookmarks=true,breaklinks=true,letterpaper=true,colorlinks,linkcolor=black,citecolor=blue,urlcolor=black]{hyperref}

\fancypagestyle{firstpage}{
	\fancyhf{}
	\setlength{\headheight}{50pt}
	
	\fancyhead[C]{
			} 
	\pagenumbering{arabic}
}  
\title{E-PUR: An Energy-Efficient Processing Unit for Recurrent Neural Networks}
\author{Franyell Silfa, Gem Dot, Jose-Maria Arnau, Antonio Gonzalez\\
Computer Architecture Deparment, Universitat Politecnica de Catalunya\\
\{fsilfa, gdot, jarnau, antonio\}@ac.upc.edu
}
\begin{document}
\maketitle
\thispagestyle{firstpage}
\pagestyle{plain}

\begin{abstract}
Recurrent Neural Networks (RNNs) are a key technology
for emerging applications such as automatic speech
recognition, machine translation or image description. Long
Short Term Memory (LSTM) networks are the most
successful RNN implementation, as they can
learn long term dependencies to achieve high accuracy.
Unfortunately, the recurrent nature of LSTM networks
significantly constrains the amount of parallelism and,
hence, multicore CPUs and many-core GPUs exhibit poor
efficiency for RNN inference.

In this paper, we present E-PUR, an energy-efficient
processing unit tailored to the requirements of LSTM
computation. The main goal of E-PUR is to support large
recurrent neural networks for low-power mobile devices. E-PUR
provides an efficient hardware implementation of
LSTM networks that is flexible to support diverse
applications. One of its main novelties is a technique that
we call Maximizing Weight Locality (MWL), which
improves the temporal locality of the memory accesses for
fetching the synaptic weights, reducing the memory
requirements by a large extent.

Our experimental results show that E-PUR achieves real-time
performance for different LSTM networks, while
reducing energy consumption by orders of magnitude with
respect to general-purpose processors and GPUs, and it
requires a very small chip area. Compared to a modern
mobile SoC, an NVIDIA Tegra X1, E-PUR provides an
average energy reduction of 92x.
\end{abstract}

\input{introduction}

\input{background}
\input{epur}

\input{methodology}

\input{results}

\input{related_work}

\input{conclusions}

\bibliographystyle{ieeetr}
\bibliography{ref}

\end{document}

%% file: introduction.tex
\section{Introduction}

Recurrent Neural Network (RNN) are a state-of-the-art
machine learning approach that has achieved a tremendous
success for a wide variety of sequence-to-sequence application domains 
\cite{sutskever2014sequence,donahue2015long,vinyals2015show,venugopalan2015sequence,miao2015eesen}.
Unlike a feed-forward Deep Neural Network (DNN), an RNN remembers
information from previous inputs to improve accuracy.
Long Short Term Memory (LSTM) \cite{hochreiter1997long} networks represent
the preferred RNN implementation nowadays. LSTM cells
can remember useful information over a long period of
time, whereas it vanishes over time in other RNN
approaches. LSTM networks are currently used for many
sequence processing problems such as speech recognition
\cite{miao2015eesen}, machine translation \cite{sutskever2014sequence}
or language modeling \cite{sundermeyer2012lstm}.

This type of applications is of especial interest for mobile
devices such as tablets, smartphones or smartwatches. For
example, voice-based interfaces represent a more natural
human-computer interface than touchscreens and
keyboards. Unfortunately, there are several challenges that
hinder the deployment of LSTM networks in mobile devices. First,
accurate LSTM networks are typically quite large and,
therefore, they require substantial memory storage and
computational resources. Real-time LSTM evaluation
comes at a high energy cost that may not be acceptable for
many low-power devices. Second, due to its recurrent
nature, an LSTM network inference exhibits a significant
amount of sequential processing and limited parallelism
and thus it cannot be efficiently executed on multicore
CPUs or GPUs. Not surprisingly, our measurements on a
recent Tegra X1 mobile SoC show that the CPU and the
GPU do not achieve real-time performance for EESEN \cite{miao2015eesen}
and RLDRADSPR \cite{kim2017residual}, two state-of-the-art LSTM
networks for speech recognition.

A few FPGA-based LSTM network accelerators targeted to the
mobile segment have been presented in recent years \cite{chang2015recurrent,lee2016fpga}.
In these designs, high energy-efficiency is achieved by
storing all the synaptic weights in local memory since 
accesses to external DRAM memory consume more than
two orders of magnitude energy than accessing a
small on-chip buffer \cite{han2015learning}. Due to the requirement of
storing the entire LSTM network on-chip, the
aforementioned accelerators are restricted to small LSTM
models. Supporting larger LSTM networks, that provide
state-of-the-art accuracy, would require a significant
increase in local storage and main memory bandwidth
usage, which would incur in a high energy overhead.

In this paper, we present E-PUR, a processing unit for
recurrent neural networks that supports large LSTM models
and provides real-time performance with an energy consumption
amenable for mobile devices. E-PUR efficiently
implements in hardware the different components of an
LSTM cell, providing enough flexibility to support
LSTM networks for different applications. A main
challenge for E-PUR is fetching the weights from memory
in an energy-efficient manner. Storing them in
local memory is not feasible due to the large size of modern
LSTM models, which is in the order of tens or even
hundreds of Mbytes, but accessing off-chip memory is
extremely expensive from an energy point of view \cite{han2015learning}.
E-PUR makes a trade-off between local memory capacity and
external memory bandwidth to achieve low power,
providing local storage for just one LSTM layer. Figure 1
shows an LSTM network consisting of multiple LSTM
cells, arranged in several layers, which are recurrently
executed for processing the different elements in the input
sequence.

In E-PUR, weights for one LSTM layer are fetched from
external DRAM and stored in on-chip local memory.
Next, each cell on the layer is evaluated for the entire input sequence, reusing
the weights stored in local memory for every element in the
input sequence. The cost of accessing main memory is amortized
due to the large size of typical input sequences, which is in
the order of thousands of elements (e.g. audio frames). Due to the current trend
towards deeper neural networks,
E-PUR offers good scalability as the size of the on-chip local
memory is independent of the number of layers.

\begin{figure}[t!]
\centering
\includegraphics[width=3.375in]{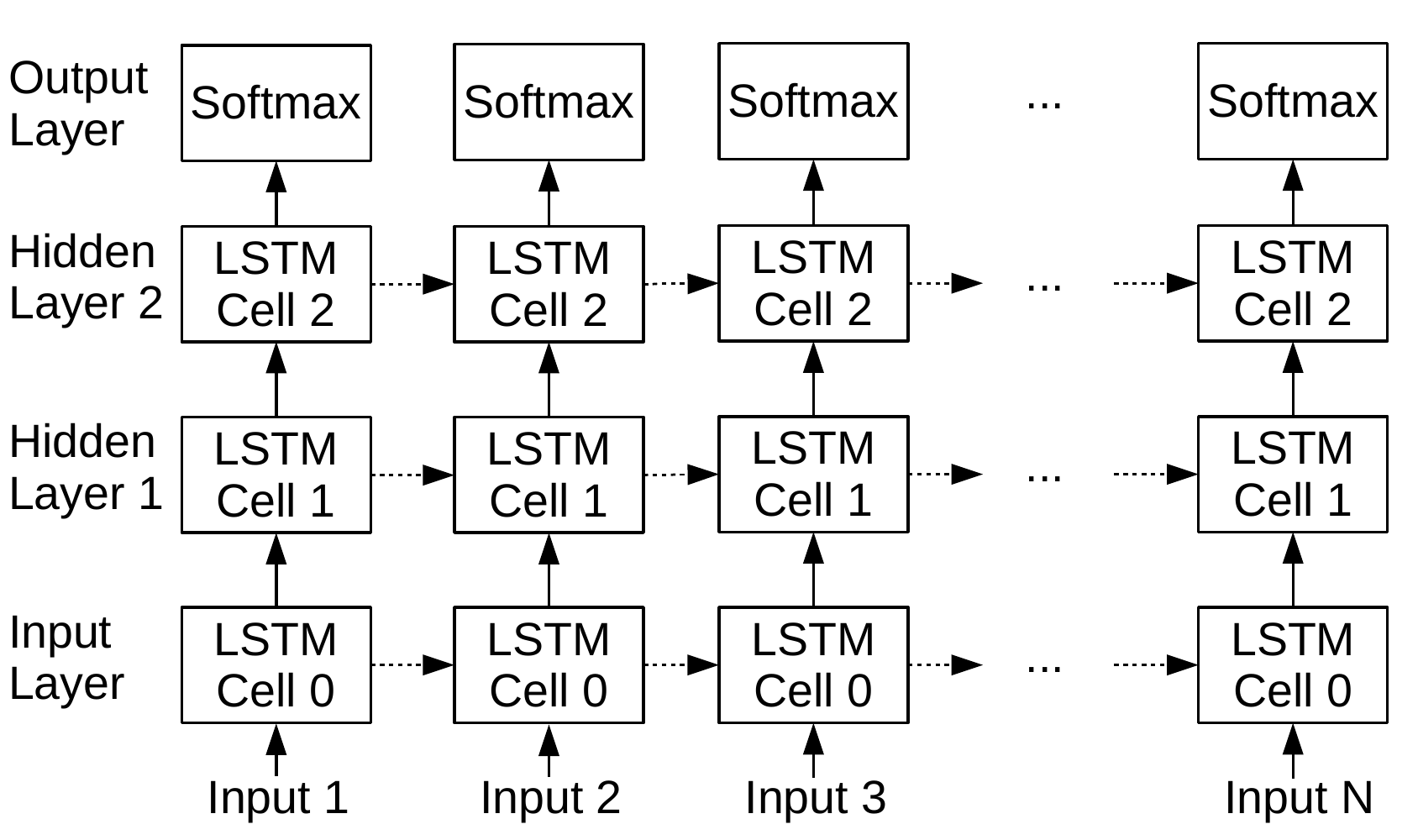}
\caption{LSTM network with three
recurrent cells and one output layer. Dotted arrows show
self-recurrent connections, a.k.a. recurrent connections,
whereas continuous arrows show connections from
previous layer, a.k.a. forward connections.}
\label{f:lstm_network}
\end{figure}

To further improve the energy-efficiency of weight
fetching, we observe that an LSTM cell has two types of
connections: self-recurrent, a.k.a. recurrent, and forward
connections from the previous layer (see Figure~\ref{f:lstm_network}).
Data dependencies impose strict sequential order for processing
recurrent connections. However, forward connections can
be processed in any order, since the results from the
previous layer are available when the current layer starts
execution. In this paper, we introduce Maximizing Weight
Locality (MWL), a technique that modifies the order in
which forward connections are processed to maximize
temporal locality. When leveraging MWL, E-PUR requires
modest local storage capacity and memory bandwidth, even
for large LSTM networks. For example, for EESEN \cite{miao2015eesen}, a speech recognition LSTM
network that has a size of 42 Mbytes, E-PUR only
requires 1.5 Mbytes of local storage. Furthermore, main
memory bandwidth usage for real-time performance is as
small as 4.2 Mbytes/s, only 0.02\% of the available memory
bandwidth of conventional low power systems such as
Tegra X1.

To summarize, this paper focuses on implementing energy-efficient,
real-time LSTM networks. Its main contributions are the following:

\begin{itemize}
\item We propose E-PUR, a processing unit for recurrent
neural networks that improves energy efficiency with
respect to CPU and GPU by orders of magnitude.

\item We introduce Maximizing Weight Locality (MWL), a
technique that dramatically improves temporal locality
of weight fetching, providing huge energy savings.

\item We evaluate E-PUR for large, representative LSTM
networks from different application domains, including
speech recognition, machine translation and video classification.

\item E-PUR achieves real-time performance while reducing
energy by 92x on average when compared with a
contemporary low power mobile SoC. Its peak
power is 975 mW and its area is 46.3 $mm^2$, which is
reasonable for most mobile devices.
\end{itemize}

The rest of the paper is organized as follows. Section~\ref{s:background}
provides some background on LSTM networks.
Section~\ref{s:epur} presents E-PUR, our processing unit for recurrent
neural networks. Section~\ref{s:methodology} describes our evaluation
methodology and Section~\ref{s:results} details the experimental
results. Section~\ref{s:related_work} reviews some related work and, finally,
Section~\ref{s:conclusions} sums up the main conclusions.

%% file: background.tex
\section{Recurrent Neural Networks}\label{s:background}

Feed-forward Deep Neural Networks (DNNs), such as
Convolutional Neural Networks (CNNs), have been shown
to be very successful for classification problems. However,
they fail to provide an effective framework for sequence-to-sequence
machine learning applications (e.g. machine
translation) for several reasons. First, the input/output
dimensionality of a feed-forward DNN is fixed, whereas
sequence processing problems require variable length
input/output. Second, DNNs use a fairly constrained
amount of context information to make a prediction,
typically a few frames of audio/video or a few words, but
some problems require taking into account distant past or
future information to be accurate. Not surprisingly,
sequence processing tasks such as machine translation or
audio/video description cannot be accurately accomplished
with the sole use of a feed-forward DNN~\cite{greff2016lstm}.
Note that a DNN can be used for a specific subtask of a sequence
processing problem, like acoustic scoring in speech
recognition, but a very expensive post processing stage is
still required to generate the output sequence~\cite{yazdani2016ultra,tabani2016ultra}.

In order to overcome the aforementioned limitations of
feed-forward DNNs, Recurrent Neural Networks (RNNs)
\cite{schuster1997bidirectional} have been proposed.
RNNs include loops or recurrent
connections, allowing information to persist from one step,
i.e. execution, of the network to the next. Therefore, RNNs
can potentially employ an unbounded amount of context
information to make predictions. In addition, RNNs are
recurrently executed for every element in the input
sequence and, hence, they can handle variable length
input/output, which is a requirement for sequence
processing problems.

Simple RNN architectures can capture and exploit short
term dependencies. However, exploiting long term
dependencies is challenging and, typically, useful
information is diluted over time in many RNN approaches.
To overcome this issue, Long Short Term Memory (LSTM)
networks were proposed \cite{hochreiter1997long}, which represent the most
successful and widely used RNN implementation, with
applications in speech recognition \cite{miao2015eesen}, machine translation
\cite{sutskever2014sequence} and language modeling \cite{sundermeyer2012lstm}.
In this section, we explain in
detail the structure and behavior of LSTM networks.

\subsection{LSTM RNN}

An LSTM RNN consists of multiple layers that are stacked together
to form a deep RNN, including an input layer and multiple
hidden layers formed by LSTM cells. These layers can be
unidirectional or bidirectional. Unidirectional layers only
use past information to perform inference for the current
execution step, whereas bidirectional layers exploit both past and
future context information and, typically, they provide
higher accuracy. Therefore, Deep Bidirectional LSTM (BiLSTM)
networks deliver state-of-the-art accuracy for multiple
sequence processing problems
\cite{baldi2001bidirectional,schuster1997bidirectional,graves2005framewise}.

\begin{figure}[t!]
\centering
\includegraphics[width=3.375in]{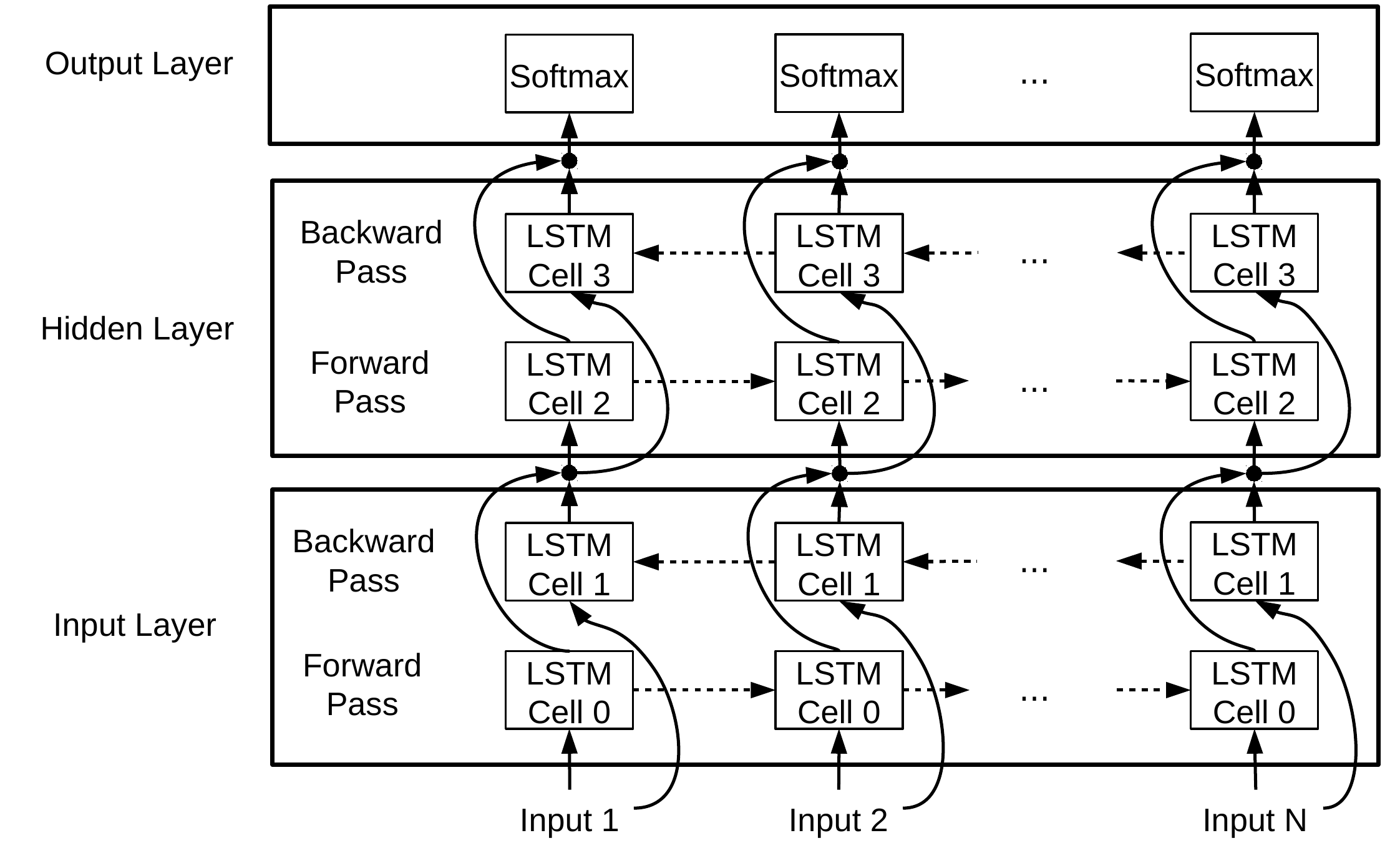}
\caption{Bidirectional LSTM network with one input
layer, one hidden layer and one output layer. Dotted arrows
show self-recurrent connections, a.k.a. recurrent
connections, whereas continuous arrows show connections
from previous layer, a.k.a. forward connections.}
\label{f:lstm_rnn}
\end{figure}

Figure~\ref{f:lstm_rnn} shows an unrolled BiLSTM network with 1
hidden layer. The bidirectional layer consists of two
LSTM cells, the first one processes the information in the
forward direction, i.e. ($x_1$) to ($x_N$), while the second one processes the input
sequence in the backward direction, i.e. ($x_N$) to ($x_1$) . Figure~\ref{f:lstm_rnn} shows
multiple instances of these two cells for each layer, which
corresponds to multiple recurrent uses of the same two cells, one
for each element in the input sequence. In this logical view
of the network, a.k.a. unrolled, recurrent connections are
shown as horizontal connections, either left-to-right or vice
versa, and they correspond in fact to connections from the
output of one cell to the input of the same cell. In a given layer, the outputs
of the  LSTM cells in both forward and backward directions are concatenated,
forming the input ($x_t$) for the next layer. Finally, a
BiLSTM network includes a feed-forward (non-recurrent)
softmax output layer, that produces the final output of the
network. For example, for speech or text applications, the
outputs represent the likelihoods of the different characters,
phonemes or words at each step.

\subsection{LSTM Cell}\label{s:lstm_cell}

\begin{figure}[t!]
\centering
\includegraphics[width=3.375in]{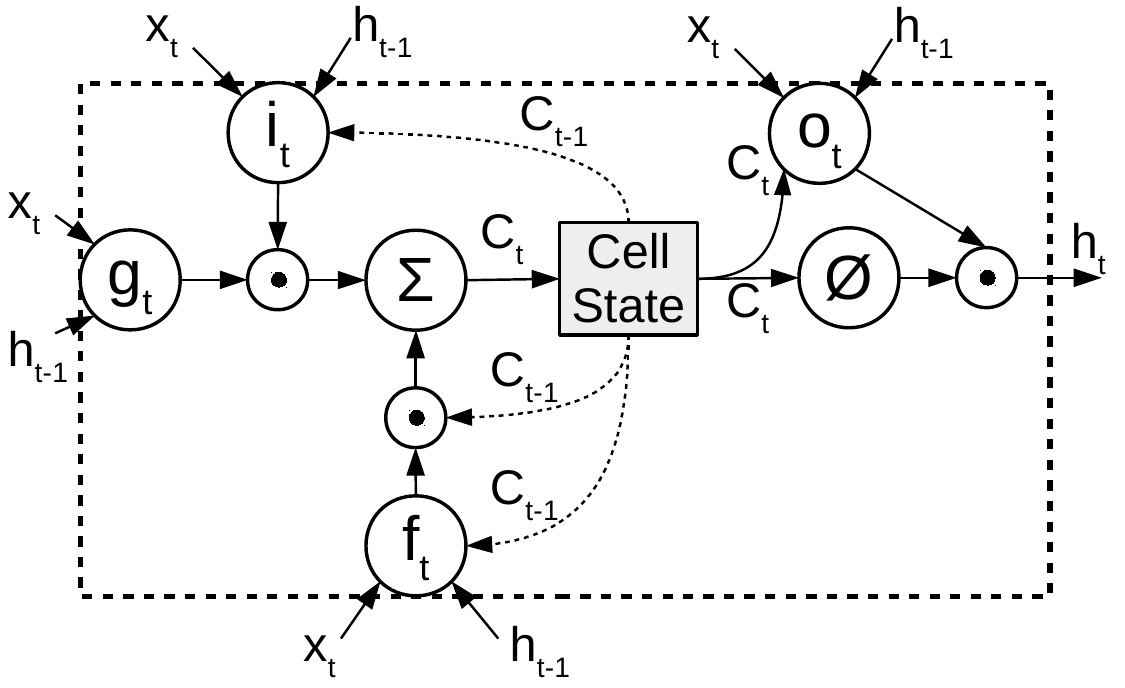}
\caption{Structure of an LSTM cell. $\odot$ denotes an element-wise 
multiplication of two vectors. $\phi$ denotes the hyperbolic
tangent. Dotted arrows represent peephole connections.}
\label{f:lstm_cell}
\end{figure}

Figure~\ref{f:lstm_cell} shows the basic structure of an LSTM cell. A key
component is the cell state ($c_t$), which represents the
memory storage of the cell. On each cell, the state is updated by
four components, commonly named as gates, which also
perform the computation of the cell output ($h_t$). Each of these
gates consists of two fully-connected networks: one taking as input
the output of the previous LSTM layer ($x_t$) and one taking as
input the output of the LSTM cell in the previous time step ($h_{t-1}$). 
The former is the one using forward connections, whereas the latter 
includes the recurrent or feedback connections.

Figure~\ref{f:lstm_equations} shows the computations performed
within an LSTM cell.
For each new element ($x_t$) of the input sequence, the following
actions are taken: First, the cell updater gate
($g_t$) modulates the amount of input information that is 
considered a candidate to update the cell state. Then, the
input gate ($i_t$) decides how much of the candidate information
will be stored into the cell state. On the other hand, the
forget gate ($f_t$) determines how much information will be
removed from the current cell state ($c_{t-1}$), i.e. which information is no
longer useful for future predictions. Finally, the output gate
($o_t$) decides the amount of information that will be
emitted from the cell.

\begin{figure}[t!]
\centering
\begin{align}
i_t = \sigma(W_{ix} x_t + W_{ih} h_{t-1} + W_{ic} \odot c_{t-1} + b_i)
\label{e:input_gate}
\end{align}
\begin{align}
f_t = \sigma(W_{fx} x_t + W_{fh} h_{t-1} + W_{fc} \odot c_{t-1} + b_f)
\label{e:forget_gate}
\end{align}
\begin{align}
g_t = \phi(W_{gx} x_t + W_{gh} h_{t-1} + b_g)
\label{e:update_gate}
\end{align}
\begin{align}
c_t = f_t \odot c_{t-1} + i_t \odot g_t
\label{e:cell_state}
\end{align}
\begin{align}
o_t = \sigma(W_{ox} x_t + W_{oh} h_{t-1} + W_{oc} \odot c_t + b_o)
\label{e:output_gate}
\end{align}
\begin{align}
h_t = o_t \odot \phi(c_t)
\label{e:cell_output}
\end{align}
\caption{Computations of a LSTM cell. $\odot$, $\phi$, and $\sigma$ denote element-wise multiplication,
hyperbolic tangent and sigmoid function respectively.}
\label{f:lstm_equations}
\end{figure}

In other words, information that is no longer
useful is removed from the cell state using the mask
generated by the forget gate. New information is added to
the cell state applying the mask generated in the input gate to
the candidate information produced by the cell updater
gate. Then, to compute the cell output, a hyperbolic
tangent is applied to the current cell state and the resulting
value is multiplied by the mask generated in
the output gate. Therefore, the cell output ($h_t$) is a filtered
version of the cell state.

The mathematical computations performed in the four gates
are very similar, as can be seen in equations \ref{e:input_gate},
\ref{e:forget_gate}, \ref{e:update_gate}, and \ref{e:output_gate}
in Figure~\ref{f:lstm_equations}. Note that conceptually each of the four gates is composed of
multiple neurons and, as shown in Figure~\ref{f:lstm_equations}, each of them consist of
two independent feed-forward fully connected networks,
which are implemented as two matrix-vector
multiplications. Therefore, for each neuron in the four gates and in all cells, two
dot-product operations are performed: one for forward connections and one for recurrent
connections. Then, the outputs of these connections are added
to a bias ($b$) and to a peephole connection. Note that peephole connections are a masked version of the cell state and 
they are used to link the cell state to the gates. Therefore, they allow
the cell state to have control over which information is
added, removed or outputted, improving prediction
accuracy for machine learning applications that require
precise timing~\cite{gers2000recurrent}. These connections are shown as the
dotted lines in Figure~\ref{f:lstm_cell}. Finally, an activation
function is applied to the result to obtain the output value of each neuron.

%% file: epur.tex
\section{E-PUR Processing Unit}\label{s:epur}

In this section, we present E-PUR, an energy-efficient
processing unit for large LSTM networks. First, we
describe the main drawbacks of state-of-the-art solutions
for LSTM inference. Next, we present the architecture of
E-PUR, which is an energy-efficient hardware
implementation of an LSTM cell. We detail the main
parameters and trade-offs made during the design of E-PUR.
Finally, we present Maximizing Weight Locality (MWL), a
technique that largely improves the temporal locality of the
memory accesses for fetching the synaptic weights.

\subsection{Motivation}\label{s:epur_motivation}
\begin{figure}[t!]
\centering
\includegraphics[width=3.375in]{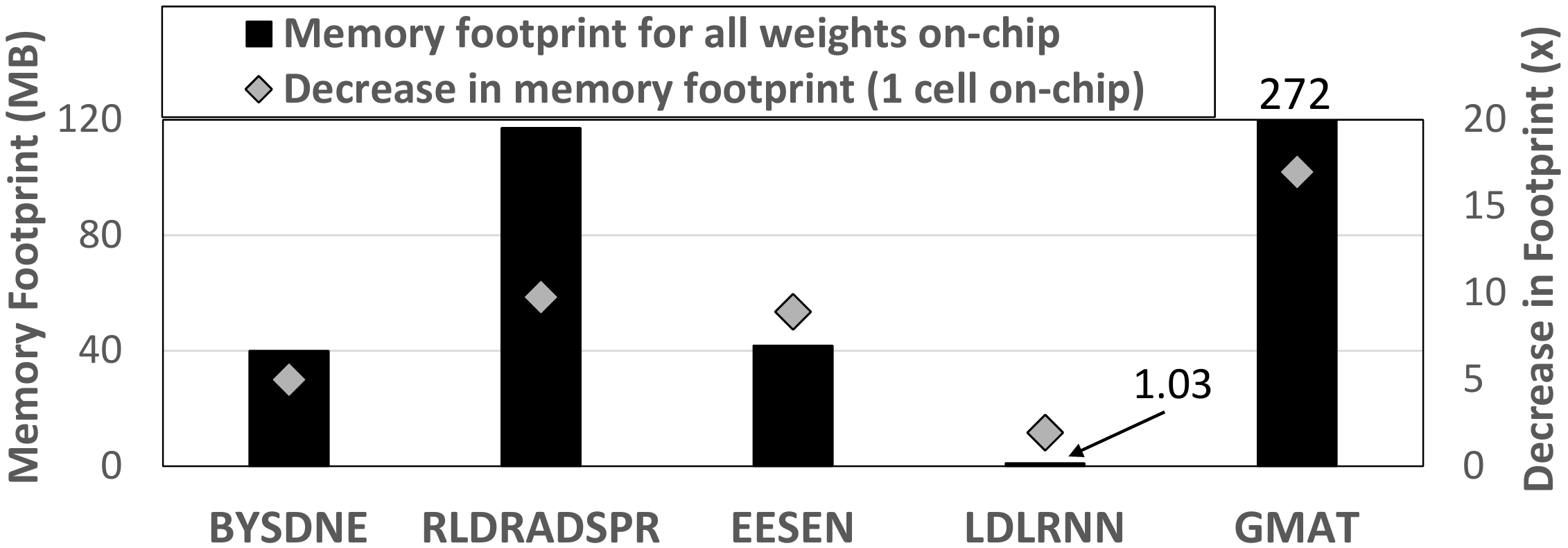}
\caption{Amount of memory required to store the synaptic
weights on-chip for several LSTM networks. Right y-axis shows the reduction in storage requirements obtained by keeping a single cell on-chip.
 }
\label{f:memory_size}
\end{figure}

State-of-the-art hardware implementations \cite{guan2017fpga,han2017ese} for LSTM networks rely on
storing all synaptic weights on-chip in order to
avoid expensive off-chip memory accesses. As
we can see in Figure~\ref{f:memory_size}, this approach is unfeasible for many
LSTM applications, due to their large memory
requirements to achieve high accuracy. For example, the
GMAT~\cite{wu2016google} LSTM network for machine translation
requires more than 250 Mbytes of memory. 

Based on the recurrent nature of LSTM networks, we propose a cost-effective
tradeoff between main memory accesses and on-chip memory storage. 
It is based on the observation that 
the input sequences of LSTM networks tend to
contain a large number of elements and for
evaluating a single pass (backward or forward) of a
given layer, only the weights for that particular layer
are used to evaluate the whole input sequence. We exploit this
characteristic of RNNs to design the memory system of E-PUR,
providing on-chip memory capacity to store only the
weights of a single LSTM layer. Note that, as seen in Figure~\ref{f:memory_size}, the storage requirements are
reduced by 7x on average, although this comes at the
expense of higher off-chip memory traffic, nonetheless this trade-off is necessary in order to support larger and deeper models
since keeping them on-chip is unfeasible due to their large memory footprint.

\subsection{Overview}\label{s:epur_overview}

\begin{figure}[t!]
\centering
\includegraphics[width=3.375in]{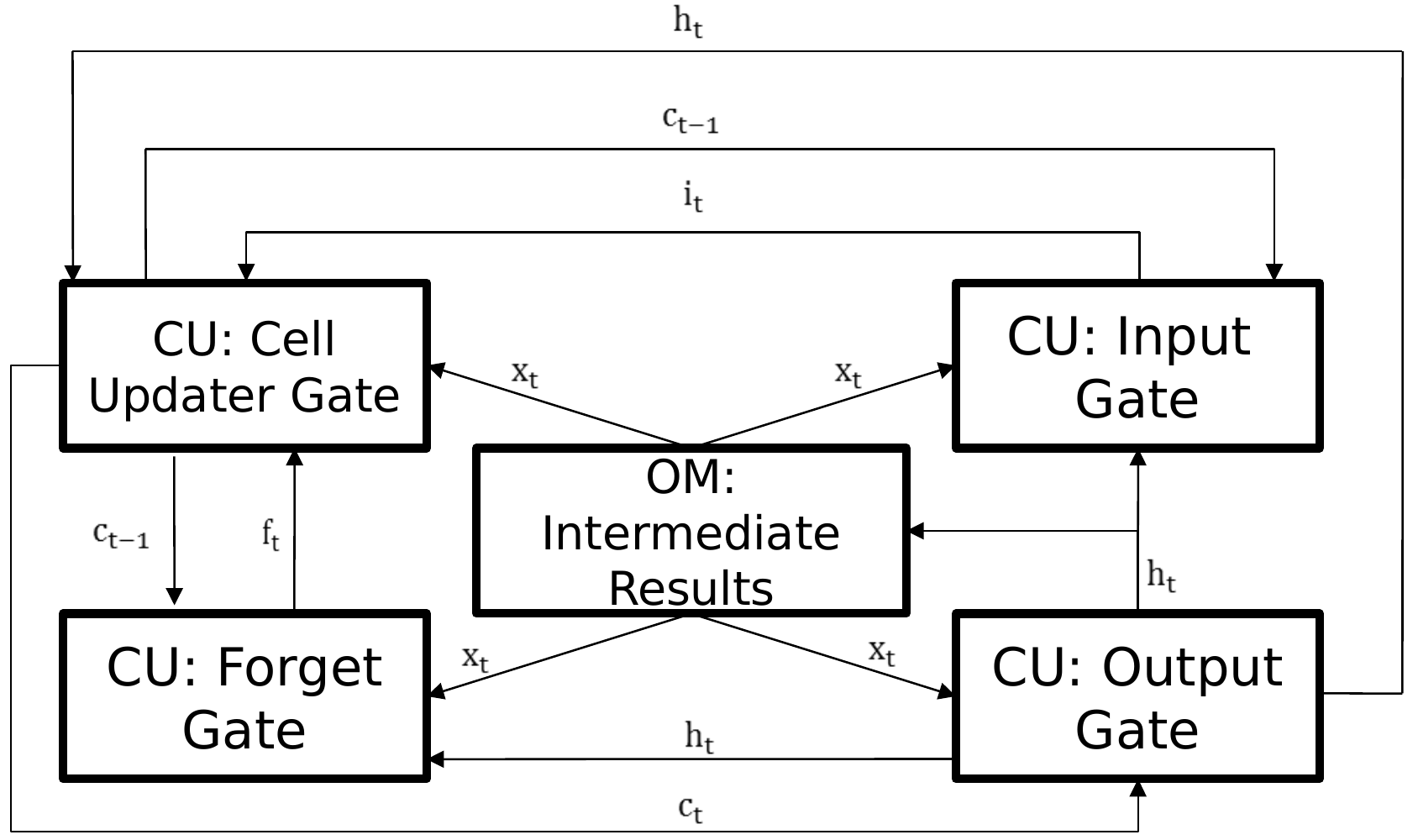}
\caption{Overview of E-PUR architecture, which consists of 4 
	computation units (CU) and an on-chip memory (OM).}
\label{f:epur_overview}
\end{figure}

Figure~\ref{f:epur_overview} shows the main components of E-PUR processing
unit. E-PUR is composed of four computation units (CUs), which have
several communications links among them. Each of these
four hardware units is tailored to the computation of one of
the four LSTM gates (i.e., forget gate, input gate, cell
updater gate and output gate). The reason for this one-to-one gate-to-CU mapping is 
that exchanging information between LSTM gates 
is not needed for most of the cell state computation.

The computation on a gate is mainly dominated by the calculation of 
the matrix-vector multiplications detailed in section~\ref{s:lstm_cell}.
Note that each gate performs exactly two matrix-vector multiplications (i.e. two dot products for each neuron) per element of the input sequence 
and, therefore, the total computation is well balanced among the
four gates. However, a minimal amount of information is shared
among CUs at the end of the cell state calculation, in order 
to gather the necessary data for its update.
As shown in Figure~\ref{f:epur_overview}, both input and forget gates send their result to
the cell updater gate, whereas the result produced in the cell updater gate is consumed
by the output gate. Moreover, after the cell state is updated by the cell updater gate, 
it is sent to the input and forget gates.

\begin{figure}[t!]
\centering
\includegraphics[width=3.375in]{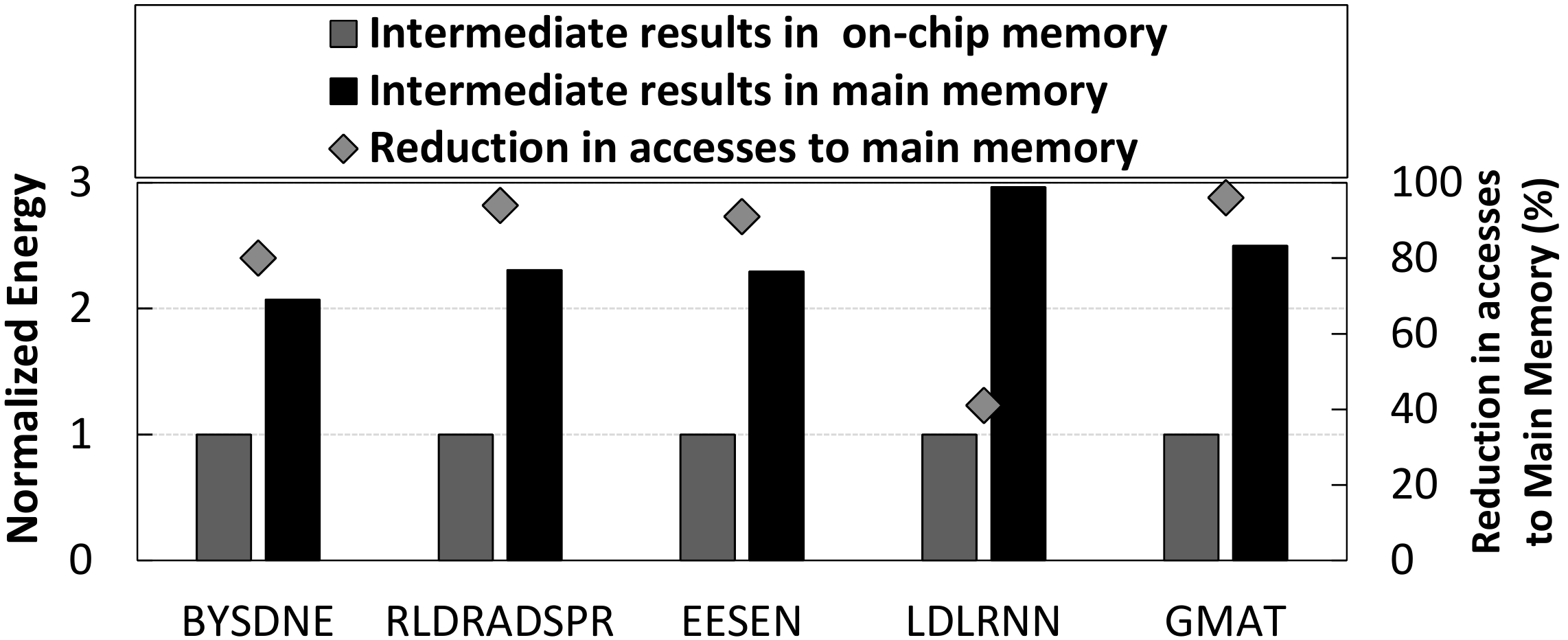}
\caption{Total energy consumption by storing intermediate
results in on-chip memory versus main memory. Right y-axis
shows the reduction in accesses to main memory.}
\label{f:normalized_energy}
\end{figure}

On the other hand, because of multiple data dependencies, the intermediate results 
produced by one layer for an entire input sequence must be saved in memory. There 
are two main alternatives to store this information: a dedicated on-chip
memory (OM) or main memory. In Figure~\ref{f:normalized_energy}, we
show the normalized energy consumption and the reduction in accesses to main memory
 for some of the most
common LSTM applications using both approaches. As we
can observe, using a dedicated on-chip memory consumes
on average 2.4x less energy than storing/loading continuously the
intermediate results to/from main memory since, on average, 77\% of 
the accesses to main memory are avoided. Therefore, this is the adopted 
solution in E-PUR. This dedicated on-chip memory is
divided in two parts of equal size. One part is used
to store the output results produced in the current layer and
the other one is used to read the results produced in the
previous layer. The reason for this double buffering is that
any result from the previous layer cannot be overwritten until the complete input
sequence has been evaluated.
\vfill\eject

\subsection{Computation Unit}\label{s:computation_unit}

\begin{figure}[t!]
\centering
\includegraphics[width=3.375in]{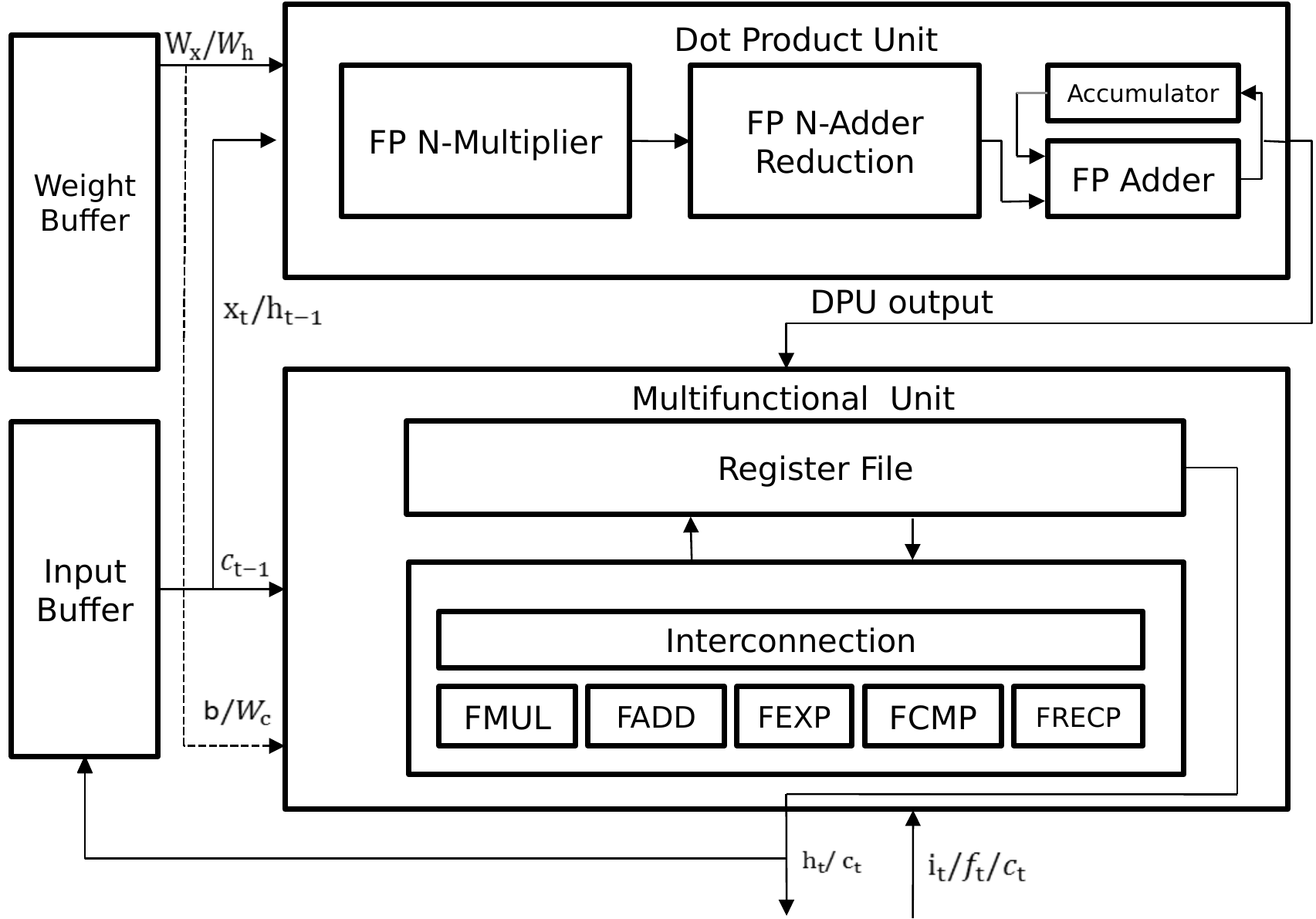}
\caption{Structure of Computation Unit.}
\label{f:computation_unit}
\end{figure}

The Computation Unit is the hardware structure that implements 
the formal model of an LSTM cell, described in Figure~\ref{f:lstm_equations}. It is composed of two main
components: the Dot Product Unit (DPU) and the
Multifunctional Unit (MU). The DPU, shown at the top
of Figure~\ref{f:computation_unit}, performs the necessary dot product operations in a gate, which is the most time-consuming part. Note that our design employs dot products over matrix-matrix multiplications 
to simplify the hardware. 
The MU, shown at the bottom of Figure~\ref{f:computation_unit},
performs the rest of operations, such as activation functions or peephole calculations.
In addition to these components, two memory buffers are used to store the input sequence and the
synaptic weights for each gate in the LSTM cell. Note that the same weights are reused for each 
recurrent execution of an LSTM cell.  

\subsubsection{The Dot Product Unit}

The DPU performs a floating point (FP) dot product 
between two vectors of length \textit{M} by splitting them into  \textit{K} sub-vectors of size  \textit{N}.
On each cycle, this unit executes the following steps. First, two size \textit{N}  sub-vectors are loaded from two
different on-chip scratchpad memories: the Weight Buffer
and the Input Buffer. The former keeps all the synaptic
weights of a given layer. The
latter stores either the input vector \textbf{$x_t$} or the previous output vector \textbf{$h_{t-1}$} of the layer being evaluated. Next, the  N-element FP Multiplier 
performs an element-wise multiplication of the two sub-vectors.
Then, the resulting vector is sent to
the  N-element FP Reduction Adder, in order to 
sum together all its elements, which takes $\log_2(N)$
cycles. Finally, the resulting value is added to the value
stored in a register called Accumulator, which accumulates 
the partial dot product until the results of all \textit{K} sub-vectors are
added together.

As shown in Figure~\ref{f:lstm_equations}, to evaluate a neuron in a given gate, two
 dot product operations are required; one takes $x_t$ as
input vector and the other one takes $h_{t-1}$. The
resulting output values of these two operations are added.
In the Computation Unit, these two dot product operations
are computed sequentially for each neuron, so that
the latter is automatically added to the result of the former
in the Accumulator register. Then, the resulting value is
sent to the Multifunctional Unit (MU), which performs the
remaining operations depending on the gate. Note that
when a value is sent to the MU, the DPU does not wait
until the MU finishes. Instead, it proceeds with the
evaluation of the remaining neurons since they
do not depend on the previous ones.

\subsubsection{The Multifunctional Unit}

The Multifunctional Unit (MU) is a configurable hardware
component whose activity depends on the Computation
Unit (i.e. input gate) where it is located, and the configuration provided by
the user. One input to the MU is the DPU output value, which corresponds to neuron's evaluation for forward and recurrent connections. On the other
hand, some of the operations performed in a particular MU
may require values produced in other Computation Units,
as explained in section~\ref{s:epur_overview}.

As shown in Figure~\ref{f:computation_unit}, an MU is composed of a register file,
an interconnection network and several floating point
units that implement basic operations:
multiplication, addition, division, comparison and
exponential. Also, each MU receives the required synaptic
information, weights for peephole connections and
biases, through the Weight Buffer. Moreover, the
previous cell state (i.e. $c_{t-1}$ for the
previous element in the input sequence) comes through the
Input Buffer.

In Table~\ref{t:mu_steps}, we detail the basic steps performed by the four
MUs once the output data from the DPUs is available. For
the sake of simplicity, we assume a single cycle per
operation and data transfer in Table~\ref{t:mu_steps}.
Note that for the evaluation we use Synopsys Design Compiler to set
realistic latencies for the different operations and
data transfers, as reported in Table~\ref{t:epur_params}. 
MUs are not in the critical path, since the DPU operations are more time
consuming and, thus, there is slack to accommodate multi-cycle latencies
for MU operations.

The MUs for the input and forget gates perform very similar
operations: they perform the multiplications for peephole
connections and add the bias. Next, they apply the sigmoid
function to the result. After this, the resulting value is sent to
the MU of the cell updater gate, which uses this
information to proceed with the computation of the 
cell state, i.e. $c_t$, and, then, it applies the hyperbolic tangent
function to this value. Once this information is
computed, it is sent to the MU of the output gate, which
computes the $k^{th}$ element of the output vector, i.e. $h_t$,
corresponding to the current element of the input sequence (i.e. $x_t$).
Finally, this value is sent to the Input Buffer of all the Computation Units. 
In addition, it is sent to the dedicated on-chip memory where it is stored to 
be consumed by the next layer, as described in Section~\ref{s:epur_overview}.
Communication between MUs is performed by dedicated links, as shown in 
Figure~\ref{f:epur_overview}. 

\newcommand{\pluseq}{\mathrel{+}=}
\begin{table*}[t!]
\caption{Steps of the Multifunctional Units for a single data element.}
\label{t:mu_steps}
\centering
\begin{adjustwidth}{-1cm}{}
\begin{tabular}{|c|c|c|c|c|c|c|c|c|c|}
\hline
\cellcolor[gray]{0.85}\textbf{stage} &\cellcolor[gray]{0.85} \textbf{0} &\cellcolor[gray]{0.85} \textbf{1} &\cellcolor[gray]{0.85} \textbf{2} &\cellcolor[gray]{0.85} \textbf{3} &\cellcolor[gray]{0.85} \textbf{4} &\cellcolor[gray]{0.85} \textbf{5} &\cellcolor[gray]{0.85} \textbf{6} &\cellcolor[gray]{0.85} \textbf{7} &\cellcolor[gray]{0.85} \textbf{8}\\
\hline
\small\textbf{Input}& \small$R_0 = DPU_O$ & & & \multicolumn{4}{c|}{\cellcolor[gray]{0.95}Sigmoid function} & &\\ 
\small\textbf{Gate}&\small$R_1 = W_{ic} \odot c_{t-1}$&\small$R_0 \pluseq R_1$&\small$R_0 \pluseq b_i$&\small$R_0=-R_0$&\small$R_0=e^{R_0}$&\small$R_0 \pluseq 1$&\small$R_0=\frac{1}{R_0}$&\small send $i_t$&\\\hline

\small\textbf{Forget}& \small$R_0 = DPU_O$ & & & \multicolumn{4}{c|}{\cellcolor[gray]{0.95}Sigmoid function} & &\\ 
\small\textbf{Gate}&\small$R_1 = W_{fc} \odot c_{t-1}$&\small$R_0 \pluseq R_1$&\small$R_0 \pluseq b_f$&\small$R_0=-R_0$&\small$R_0=e^{R_0}$&\small$R_0 \pluseq 1$&\small$R_0=\frac{1}{R_0}$&\small send $f_t$&\\\hline

\small\textbf{Cell}& &\multicolumn{4}{c|}{\cellcolor[gray]{0.95}Hyperbolic tangent function}& & & & \\ 
\small\textbf{Updater}&\small$R_0 = DPU_{O} + b_c$&\small$R_1 = -R_0$&\small$R_1 = e^{R_1}$&\small$R_1=R_0-R_1$&\small$R_0=\frac{R_1}{R_0}$&\small wait&\small wait&\small recv&\small$R_0=R_0*i_t$\\
\small\textbf{Gate}& &\small$R_0 = e^{R_0}$& &\small$R_0=R_0+R_1$& &\small$i_t \& f_t$&\small$i_t \& f_t$&\small $i_t \& f_t$&\small$R_1=f_t*c_{t-1}$\\\hline

\small\textbf{Output}& & & & & & & & &\\
\small\textbf{Gate}&\small$R_0 = DPU_O + b_o$&\small wait $c_t$&\small wait $c_t$&\small wait $c_t$&\small wait $c_t$&\small wait $c_t$&\small wait $c_t$&\small wait $c_t$&\small wait $c_t$\\\hline

\hline\hline

\cellcolor[gray]{0.85}\textbf{stage} &\cellcolor[gray]{0.85} \textbf{9} &\cellcolor[gray]{0.85} \textbf{10} &\cellcolor[gray]{0.85} \textbf{11} &\cellcolor[gray]{0.85} \textbf{12} &\cellcolor[gray]{0.85} \textbf{13} &\cellcolor[gray]{0.85} \textbf{14} &\cellcolor[gray]{0.85} \textbf{15} &\cellcolor[gray]{0.85} \textbf{16} &\cellcolor[gray]{0.85} \textbf{17}\\\hline

\small\textbf{Cell}& &\multicolumn{4}{c|}{\cellcolor[gray]{0.95}Hyperbolic tangent function}& & & & \\ 
\small\textbf{Updater}&\small$R_0 = R_0 + R_1$&\small$R_1 = -R_0$&\small$R_1 = e^{R_1}$&\small$R_1=R_0-R_1$&\small$R_0=\frac{R_1}{R_0}$&\small send $\phi(c_t)$& & & \\
\small\textbf{Gate}& &\small$R_0 = e^{R_0}$& &\small$R_0=R_0+R_1$& & & & & \\
& & & & & & & & &\\\hline

\small\textbf{Output}& & &\small$R_1 =$ & &\multicolumn{4}{c|}{\cellcolor[gray]{0.95}Hyperbolic tangent function}  &\\
\small\textbf{Gate}&\small wait $c_t$&\small recv $c_t$&\small $W_{oc} \odot c_t$&\small $R_0 = R_0 + R_1$&\small $R_0=-R_0$&\small $R_0=e^{R_0}$&\small $R_0 \pluseq 1$&\small $R_0=\frac{1}{R_0}$&\small $h_t=R_0*R_1$\\
 & & & & & &\small recv $\phi(c_t)$&\small$R_1=\phi(c_t)$ & &\\\hline

\end{tabular}
\end{adjustwidth}
\end{table*}

\subsection{MWL: Maximizing Weight Locality}\label{s:MWL}

\begin{figure}[t!]
\centering
\includegraphics[width=3.375in]{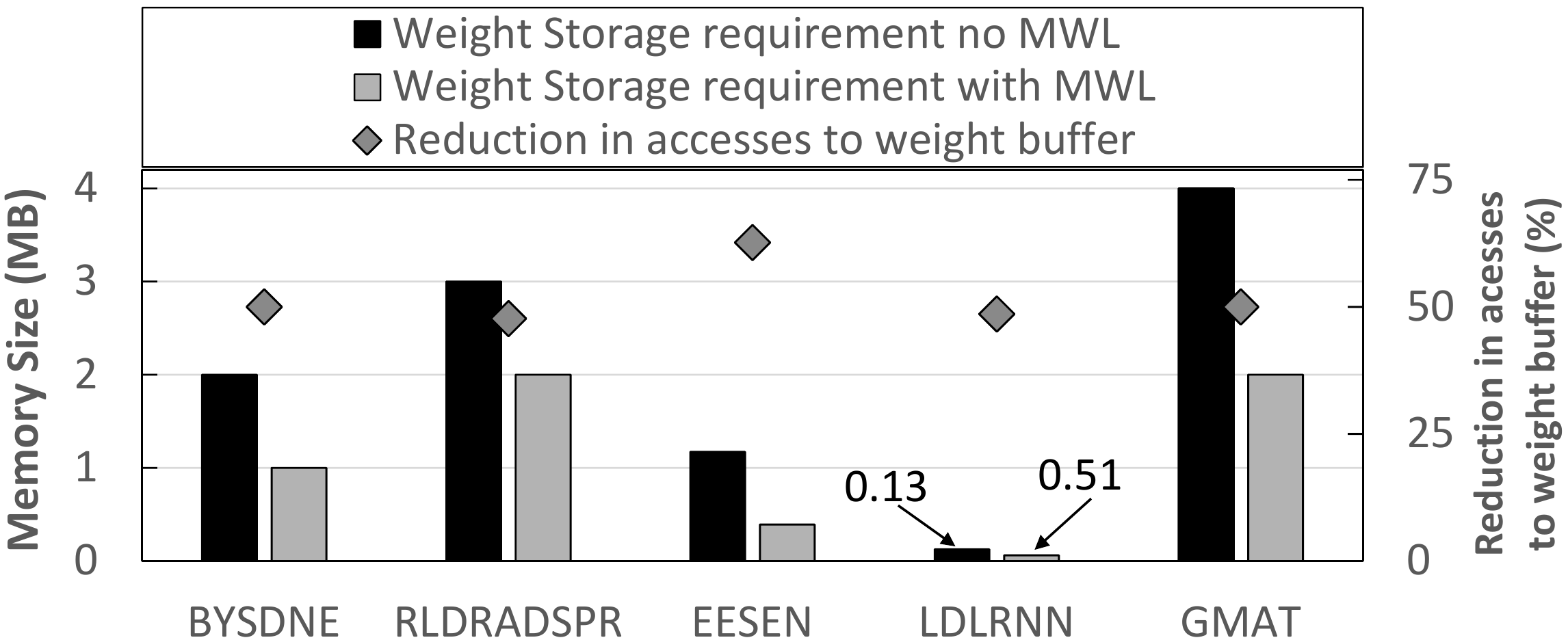}
\caption{Synaptic weights memory requirements for a
single LSTM cell with and without MWL. Right y-axis shows the reduction in accesses to the weight buffer.}
\label{f:memory_size_cell}
\end{figure}

As shown in Figure~\ref{f:memory_size_cell}, on-chip memory requirements 
to store the synaptic weights are still quite significant for some 
applications (i.e. GMAT), despite the optimizations proposed in Section~\ref{s:epur_motivation}. 
In order to further improve energy consumption and reduce on-chip memory requirements, 
we propose a technique that maximizes temporal locality of the accesses to the
weights, which are performed for each layer. We call
this technique Maximizing Weight Locality (MWL). The key observation
is that forward connections (i.e. their
inputs come from the previous layer) can be processed in
any order since all the output results from the previous
layer are available. Therefore, E-PUR processes forward
connections in an order that improves temporal locality.
The idea is that in a given gate, instead of completely evaluating all the neurons for
a single element ($x_t$) of the input sequence, the evaluation for all the neurons is split in two steps. In
the fist step, all the neurons are evaluated using as input the forward connections for
the whole input sequence (i.e, $x_t$, .., $x_n$) and the intermediate results are saved. For the second step,
MWL proceeds with the computation of all neurons for the recurrent connections (i.e, $h_{t-1}$, .., $h_{n-1}$).
Note that in this case, the evaluation must be done in sequence since data dependencies in the recurrent connections impose strict sequential
order.

With this approach, E-PUR reuses 
a small subset of the weights, those corresponding to a particular neuron, at extremely short distances. Note that for a given neuron, 
once it is partially computed for all elements of the 
input sequence, its corresponding weights will no longer be required 
and, thus, they can be evicted from on-chip memory. Therefore, while 
processing forward connections, E-PUR only requires on-chip storage
for the forward weights of a single neuron at a time, significantly reducing 
on-chip storage requirements and energy consumption. As shown in Figure~\ref{f:memory_size_cell}, 
the storage requirements for the weights are reduced by approximately 50\% on average. Note that
recurrent connections are evaluated as
usual and, hence, all the associated weights for a given 
layer must be stored on-chip to avoid excessive accesses to off-chip memory.  

The drawback of MWL is that requires additional
memory to store the partial evaluations of all neurons on a given layer. In the
design of E-PUR, presented in Section~\ref{s:computation_unit},
 neurons in a cell are completely evaluated for an element in the input sequence
before proceeding to the next input element. Therefore, only the
final output vector of a cell, $h_t$, has to be stored in a memory buffer. On the other hand, with MWL, the neurons are first partially evaluated for all the elements in the input sequence,
by operating exclusively on the forward connections. In
this case, the partial evaluations for the neurons in each of the four gates
must be stored, since later they have to be merged with the
result of evaluating the recurrent connections, in order to produce the final
output. This requires an increase in on-chip
storage requirements for intermediate results, but this overhead is minimized applying linear quantization to
the partial output results. Next subsections provide further details
on the implementation and trade-offs of MWL.

\subsubsection{Prioritize Forward Connections}\label{s:prioratize_forward}

\begin{figure}[t!]
\centering
\includegraphics[width=3.375in]{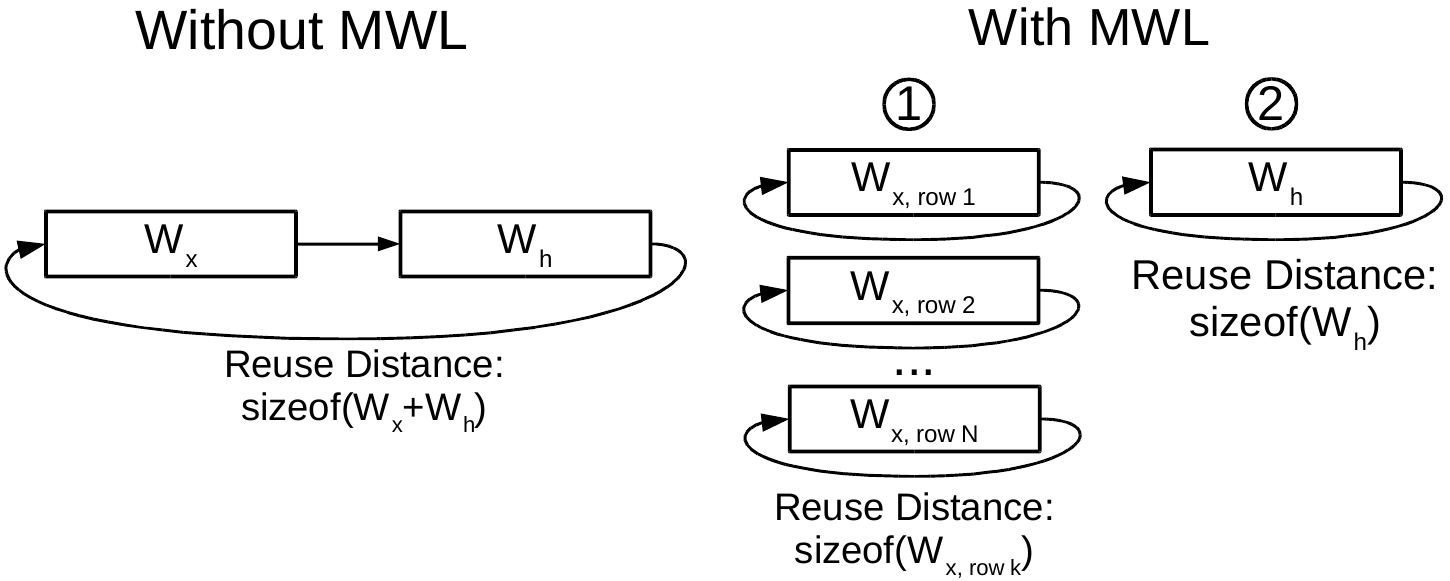}
\caption{Reuse distance for the accesses to the weight
information.}
\label{f:mwl_reuse_distance}
\end{figure}

The conventional way to evaluate the input
sequence in a layer is by performing all the necessary computations of
the current element in the input sequence before starting with the
next one. It implies the evaluation of both forward and recurrent connections
 in each layer. However, by following
this order, the temporal locality to access the weights
from each gate is suboptimal. As we can see in the left part of
Figure~\ref{f:mwl_reuse_distance}, the reuse distance of a weight access is
equal to adding the size of the two weight matrices, i.e. $W_x$ and $W_h$.
This has a direct impact on storage requirements, since a longer 
reuse distance requires a larger on-chip memory to hold the weights in
order to avoid expensive off-chip memory accesses.

MWL improves temporal locality in the weight accesses by changing the
evaluation order of the two feed-forward networks across the
entire input sequence in a given layer. It is based on the
observation that all feed-forward networks that take $x_t$
as input vector, i.e. those that contain forward
connections, do not depend on the previous output of the layer,
as we can see in Figure~\ref{f:lstm_equations}.
Therefore, we improve temporal locality by partially evaluating all the neurons in
a layer for the entire input sequence and then proceeding
with the recurrent connections ($h_{t-1}$), instead of sequentially evaluating the neurons in the layer for
$x_t$ and $h_{t-1}$ and then proceeding with $x_{t+1}$ and $h_{t}$. This 
reduces the storage requirements to the size of a single feed-forward
network, as seen in Figure~\ref{f:memory_size_cell}.

Note that for a given neuron in a cell, its 
computations use the same subset of weights (i.e, a single row from the 
weight matrix of the feed-forward network), therefore the reuse distance is reduced
to a single row of the feed-forward matrix, as we can see in the middle part of
Figure~\ref{f:mwl_reuse_distance}. Henceforth, we store them in a small buffer (i.e. 4KB), thus, avoiding 
to access the weight buffer for the forward connections. As a result, as shown in 
Figure~\ref{f:memory_size_cell}, the accesses to
the weight buffer are reduced by 50\% on average.

Finally, after the partial evaluation of the forward connections for all the neurons in a 
layer, the evaluation for recurrent connections 
is performed as explained in Section~\ref{s:epur_overview}, i.e. the next input is not evaluated until the results of the current input are computed, 
to respect data dependencies (right part of Figure~\ref{f:mwl_reuse_distance}).

\subsubsection{Storage of the Intermediate Results}

The dedicated on-chip memory for intermediate results
(see Section~\ref{s:epur_overview}) is dimensioned to hold the final outputs
(i.e. $h_t$) for a given layer, which are produced by the
output gates in each cell. When using MWL, the temporal values
produced by each gate while evaluating forward
connections must be saved for the entire
input sequence since the MUs  will need these values to
compute the final outputs, as
explained above. Therefore, the main drawback of this
technique is the extra storage requirements for these
intermediate values, which is equal to four times the
memory needed to store the $h_t$ outputs, because
intermediate values are produced in the four gates. In order
to deal with this issue, E-PUR applies a well-known
technique, linear quantization, which reduces the number of
bits needed to represent these values, at the expense of
potentially some loss in accuracy. More specifically, we
apply linear quantization using 8 bits
per element introducing negligible accuracy loss in our
set of neural networks. Empirically we found that for the networks EESEN and RLDRASPR the 
WER decreases by less than 1\%. For the other three networks (BYSDNE, LDLRNN, GMAT), 
we observed an accuracy loss of less than 
0.5\%. Note that previous work reported similar results~\cite{wu2016google,linTA15}.

When using linear quantization, for a given neuron $k$ with partial output (i.e. $o_k$) produced in MWL, its
quantized value (i.e. $o_k'$) is computed using the following
equations:

\begin{equation}
\beta = \frac{2^{n-1} - 1}{\alpha}
\end{equation}
\begin{equation}
o_k' = round(\beta * o_k)
\end{equation}

where $n$ is the number of bits of the quantized value
(represented as an integer), i.e. 8 bits, and $\alpha$ is the
maximum value of $o_k$. Theoretically, the value of $\alpha$ is
unbounded; however, we empirically found that its absolute
value is normally less than 20 for recurrent neural
networks. Note that the constant $\beta$ is computed offline.

In order to compute the previous equation, we extended the
MU with functional units to support AND, OR and SHIFT
operations. We implemented the rounding operation by
adding one to the product $\beta*o_k$ followed by a sequence
of AND, OR, additions and multiplications. These
operations are performed in parallel with the computation
of $o_{k+1}$ done by the DPU. Once the casting is completed,
the value is stored in the on-chip memory for intermediate
results.
 
After all the partial outputs ($o_k$) for all the neurons are computed, recurrent connections
are evaluated as explained in section~\ref{s:prioratize_forward}. However,
before computing the final output for a given gate in a cell,
the previous quantized values must be converted back to
floating point numbers and added to the result of evaluating the
recurrent connections. We implemented this value conversion
through a look up table that maps the integer quantized
value to its floating point representation. Note that the size
of this table is small since $n$ is small (i.e. 8 bits in
our experiments) and it is computed offline.

%% file: methodology.tex
\section{Evaluation Methodology}\label{s:methodology}

As our set of benchmarks, we use five recent LSTM
networks which are described in Table~\ref{t:lstm_networks}. Our 
selection includes RNNs for popular applications such as speech 
recognition, machine translation or video classification. Each of 
these networks has a different number of internal layers and 
outputs, i.e. number of cells. Moreover, there are some networks 
that only perform a single pass for inference computation, i.e.
they are unidirectional, whereas two of them, EESEN and
GMAT, are bidirectional. On the other hand, we include networks
with and without peephole connections. Therefore, our selection
covers a wide range of LSTM designs with different sizes,
from small RNNs of one Mbyte to large RNNs or hundreds
of Mbytes. For each network we used the accuracy metric listed in Table~\ref{t:lstm_networks} and
the test set provided in each work. 

\begin{table*}[t!]
\caption{LSTM Networks used for the experiments.}
\label{t:lstm_networks}
\centering
\begin{tabular}{cccccccc}
\cellcolor[gray]{0.9}\small\textbf{Network}&\cellcolor[gray]{0.9}\small\textbf{App Domain}&\cellcolor[gray]{0.9}\small\textbf{Layers}&\cellcolor[gray]{0.9}\small\textbf{Neurons}&\cellcolor[gray]{0.9}\small\textbf{Passes}&\cellcolor[gray]{0.9}\small\textbf{Peephole}&\cellcolor[gray]{0.9}\small\textbf{Size (MB)}  &\cellcolor[gray]{0.9}\small\textbf{Accuracy}   \\
\small BYSDNE~\cite{yue2015beyond}&\small Video Classification&\small 5&\small 512&\small 1&\small Yes&\small 40&\small 88.6\%\\
\cellcolor[gray]{0.9}\small RLDRADSPR~\cite{kim2017residual}&\cellcolor[gray]{0.9}\small Speech Recognition&\cellcolor[gray]{0.9}\small 10&\cellcolor[gray]{0.9}\small 1024&\cellcolor[gray]{0.9}\small 1&\cellcolor[gray]{0.9}\small Yes&\cellcolor[gray]{0.9}\small 118&\cellcolor[gray]{0.9}\small 39.3 WER\\
\small EESEN~\cite{miao2015eesen}&\small Speech Recognition&\small 5&\small 320&\small 2&\small Yes&\small 42&\small 23.8 WER\\
\cellcolor[gray]{0.9}\small LDLRNN~\cite{lipton2015learning}&\cellcolor[gray]{0.9}\small Time Series &\cellcolor[gray]{0.9}\small 2&\cellcolor[gray]{0.9}\small 128&\cellcolor[gray]{0.9}\small 1&\cellcolor[gray]{0.9}\small No&\cellcolor[gray]{0.9}\small 1&\cellcolor[gray]{0.9}\small 85\%\\
\small GMAT~\cite{wu2016google}&\small Machine Translation&\small 17&\small 1024&\small 1&\small No&\small 272&\small 24.1 Bleu\\
\end{tabular}
\end{table*}

As our baseline platform, we use an NVIDA Tegra X1 SoC
~\cite{TegraX1} whose parameters are shown in Table~\ref{t:tegrax1}. Its
energy consumption has been measured by
reading the registers of the Texas Instruments INA3221
power monitor included in the Jetson TX1
development board~\cite{TegraX1}. Regarding the software
implementation of the networks, we implemented them using Keras~\cite{chollet2015keras},
a high-level neural networks API. We use the Theano~\cite{Rami2016Theano} backend to
run the LSTM networks. Theano relies on cuBLAS, a high-performance CUDA 
library, to perform matrix operations. 
Finally, we also implemented MWL in software for the Tegra X1 (Tegra X1+MWL) to analyze the
benefits of a software-only implementation. We used CUDA to implement this version and 
employed kernel fusion~\cite{wang2010kernel} to merge the processing of different gates
in one kernel, avoiding excessive number of API calls, which represent a significant 
overhead in this platform.

\begin{table}[t!]
\caption{Tegra X1 parameters.}
\label{t:tegrax1}
\centering
\begin{tabular}{cc}
\cellcolor[gray]{0.9}\small\textbf{Parameter}&\cellcolor[gray]{0.9}\small\textbf{Value}\\
\small CPU&\small 4-core ARM A-57\\
\cellcolor[gray]{0.9}\small GPU&\cellcolor[gray]{0.9}\small 256-core Maxwell GPU\\
\small Streaming Multiprocessors&\small 2 (2048 threads/proc)\\
\cellcolor[gray]{0.9}\small Technology&\cellcolor[gray]{0.9}\small 20 nm\\
\small Frequency&\small 1.0 GHz\\
\cellcolor[gray]{0.9}\small CPU L2 Cache&\cellcolor[gray]{0.9}\small 2 MB\\
\small GPU L2 Cache&\small 256 KB\\
\end{tabular}
\end{table}

To evaluate our accelerator, we have developed a cycle-accurate simulator of
E-PUR. This simulator estimates the total energy
consumption (static and dynamic) and execution time of
LSTM networks running on top of E-PUR. We
used Verilog to implement the different pipeline components
of E-PUR, and we synthesized them using the Synopsys
Design Compiler to obtain their delay and energy
consumption. We use a typical process corner with a voltage of 0.78V and average 
switching activity is used to estimate dynamic power.
We used CACTI~\cite{muralimanohar2009cacti} to estimate the delay
and energy  (static and dynamic) of on-chip memories. Finally, to estimate timing and energy
consumption of main memory we used MICRON models~\cite{Micron}.
We model 4 GB LPDDR4 DRAM.  

Regarding the clock frequency, we used the delays reported by Synopsys
Design Compiler and CACTI to set the frequency such that most hardware structures operate at
one clock cycle. In addition, we evaluated alternative frequency values in order to
minimize energy consumption. Note that many hardware components, such as floating
point multipliers, are pipelined and have latency larger than one clock cycle, as shown
in Table~\ref{t:epur_params}.   

The remaining configuration parameters of E-PUR used for our
experiments are shown in Table~\ref{t:epur_params}. We select an
energy-efficient configuration that achieves real-time
performance for all the neural networks in Table~\ref{t:lstm_networks}. Note that E-PUR is
designed to accommodate large LSTM networks and, thus,
its on-chip storage might be over-sized for the small models
used in some applications. In this case, unused memory banks to store weights and intermediate
results are power gated to reduce static power.

\begin{table}[t!]
\caption{Hardware parameters for E-PUR.}
\label{t:epur_params}
\centering
\begin{tabular}{ccc}
\cellcolor[gray]{0.9}\small\textbf{Parameter}&\cellcolor[gray]{0.9}\small\textbf{E-PUR}&\cellcolor[gray]{0.9}\small\textbf{E-PUR+MWL}\\
\small Technology&\small 28 nm&\small 28 nm\\
\cellcolor[gray]{0.9}\small Frequency&\cellcolor[gray]{0.9}\small 500 MHz&\cellcolor[gray]{0.9}\small 500 MHz\\
\small Intermediate Memory&\small 6 MB&\small 6 MB\\
\cellcolor[gray]{0.9}\small Weights Memory&\cellcolor[gray]{0.9}\small 4 MB per CU&\cellcolor[gray]{0.9}\small 2 MB per CU\\
\small Inputs Memory&\small 8 KB per CU&\small 4 KB per CU\\
\cellcolor[gray]{0.9}\small DPU Width&\cellcolor[gray]{0.9}\small 16 operations&\cellcolor[gray]{0.9}\small 16 operations\\
\small MU Operations&\multicolumn{2}{c}{\small cycles: 2 (ADD), 4 (MUL), 5 (EXP)} \\
\cellcolor[gray]{0.9}\small MU Communication&\small \cellcolor[gray]{0.9} 2 cycles &\cellcolor[gray]{0.9}\small 2 cycles\\
\small Peak Bandwidth&\small  30 GB/s &\small 30 GB/s\\

\end{tabular}
\end{table}

%% file: results.tex
\section{Experimental Results}\label{s:results}

In this section, we present the evaluation of E-PUR, our
processing unit for RNNs. The baseline configuration used
for comparison purposes is a Theano implementation
running on a mobile NVIDIA Tegra X1 platform. The
configuration labeled as E-PUR throughout this section
consists of our first design presented in Section~\ref{s:epur_overview},
whereas the configuration E-PUR+MWL includes our
technique for improving the temporal locality of the
weights described in Section~\ref{s:MWL}. First, we present the
energy reduction achieved by these two configurations with
respect to the Tegra X1. Second, the performance
improvement over the baseline is analyzed. Third, the
power consumption for each of these
configurations is shown. Fourth, we present the total area
required by E-PUR. Finally, we analyze the performance of a
software-only implementation of MWL.

\begin{figure}[t!]
	\centering
	\includegraphics[width=3.375in]{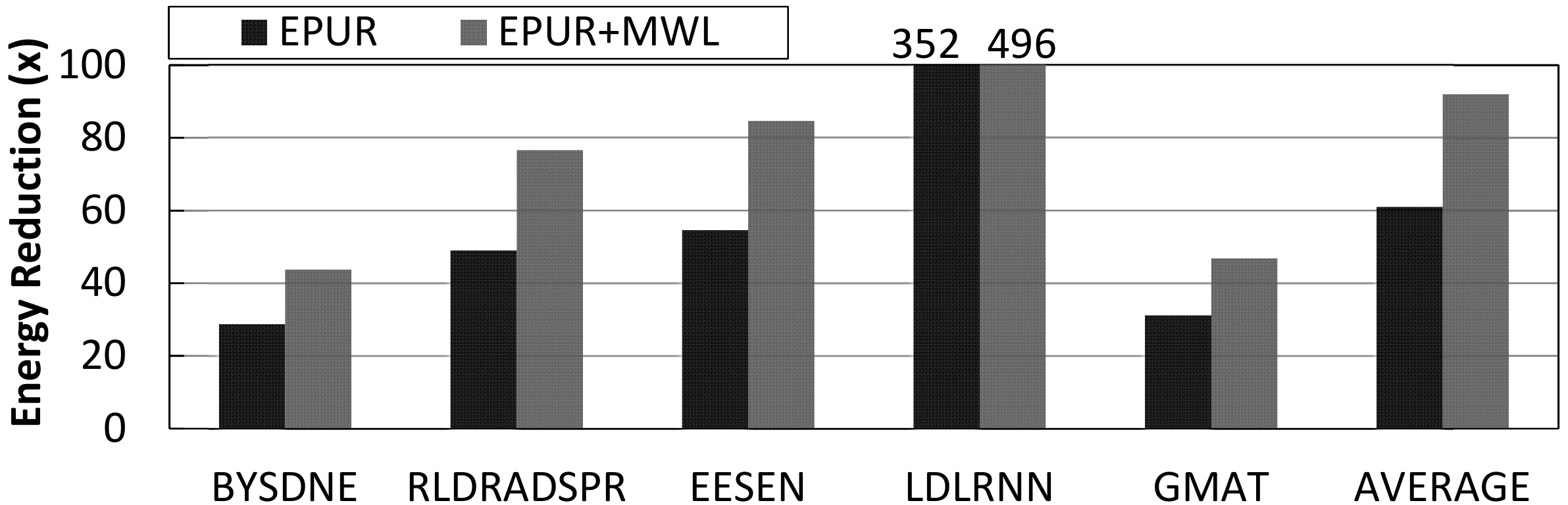}
	\caption{Energy reduction of E-PUR with respect to the Tegra X1.}
	\label{f:energy_reduction}
\end{figure}

Figure~\ref{f:energy_reduction} shows the energy reduction. On average, E-PUR
and E-PUR+MWL achieve 61x and 92x energy
reduction respectively. All the LSTM networks
show large improvements of at least 28x reduction in
energy consumption. A remarkable case is LDLRNN, 
for which E-PUR reduces the total energy
by 352.4x and 496.1x, respectively. The reason for this large energy reduction 
is that LDLRNN has fewer outputs per layer, i.e. smaller number of neurons, which means
that the matrix-vector multiplications require less number
of operations and, also, less memory accesses are done to fetch the weights or intermediate results. This penalizes Tegra X1 platform because
the ratio between computations in the GPU and other
related tasks (e.g., GPU synchronization, CPU work, etc.)
is smaller. Note that for E-PUR most of the energy savings come from avoiding accesses to
main memory to load/store intermediate results and weights. In the case of E-PUR+MWL,
 energy savings come from avoiding accesses to the on-chip memory for weights by
 50\% on average.

\begin{figure}[t!]
	\centering
	\includegraphics[width=3.375in]{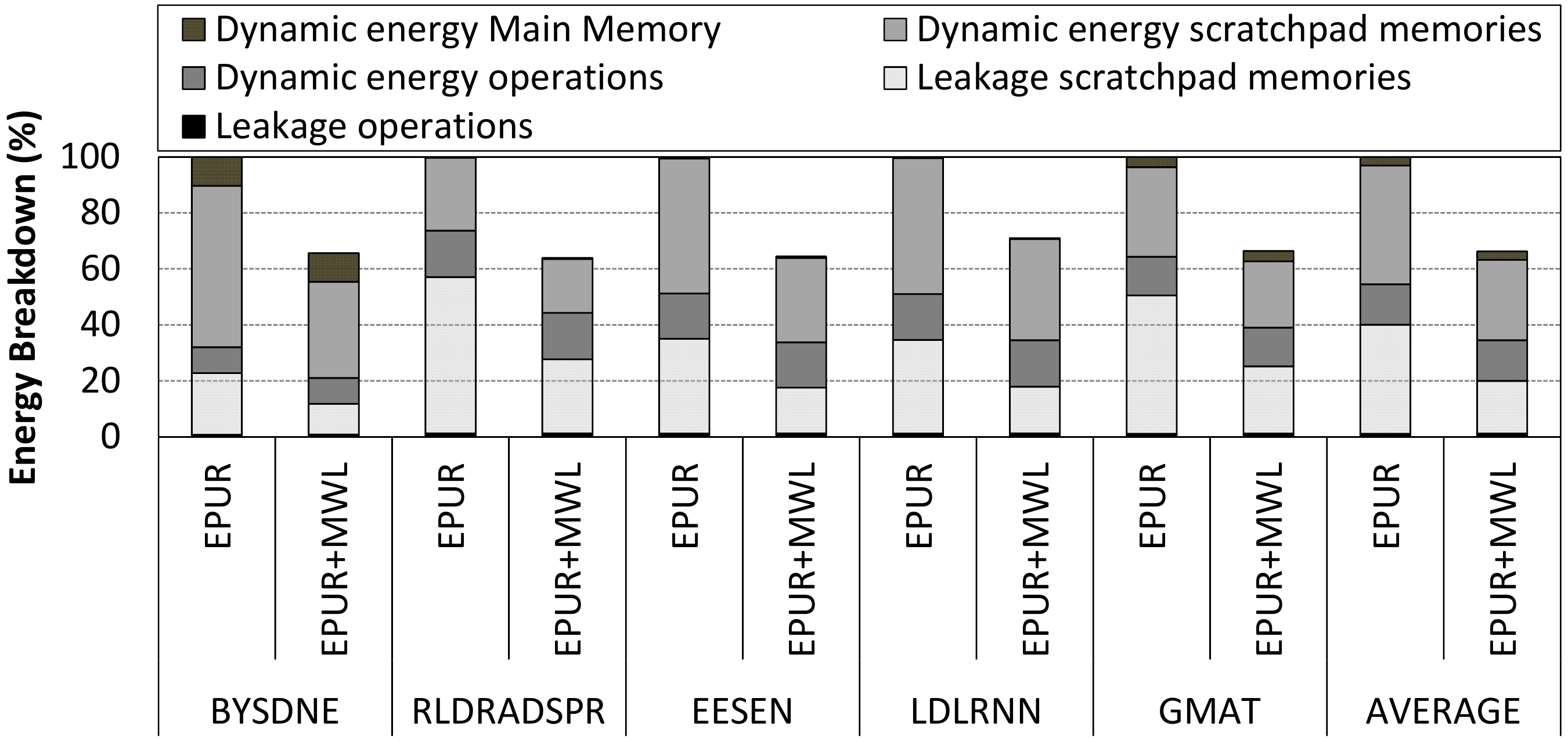}
	\caption{Energy breakdown for E-PUR and E-PUR+MWL.}
	\label{f:energy_breakdown}
\end{figure}

Figure~\ref{f:energy_breakdown} shows the energy breakdown for the two
configurations of E-PUR. The different components of E-PUR 
are grouped into ``scratchpad memories'', which
includes all the on-chip memories, and ``operations'', which
includes the pipeline components, such as the
functional units. Since on-chip memory requirements and number of memory accesses are
significant, the overall energy consumption is dominated by the dynamic accesses to on-chip 
memories, which consume around 80\%. Because MWL reduces the dynamic accesses for the weight buffer by
50\% on average, the dynamic energy due to on-chip memories is reduced in 31\% on average for E-PUR+MWL. 
Note that the energy consumption due to scratchpad memories is not reduced by 50\% since there is an 
increase in memory accesses to the on-chip memory for intermediate results.  In the case of the leakage due to on-chip memories, after applying MWL, it is reduced by more than 50\% on average. This saving comes from the reduction in storage requirements to store the weights for the forward connections. Henceforth, the savings in leakage and dynamic energy result in 35\% reduction of the total energy consumption. Regarding the energy consumption due to the
operations, it ranges between 10\% and 20\% of the total
energy for both configurations.

\begin{figure}[t!]
	\centering
	\includegraphics[width=3.375in]{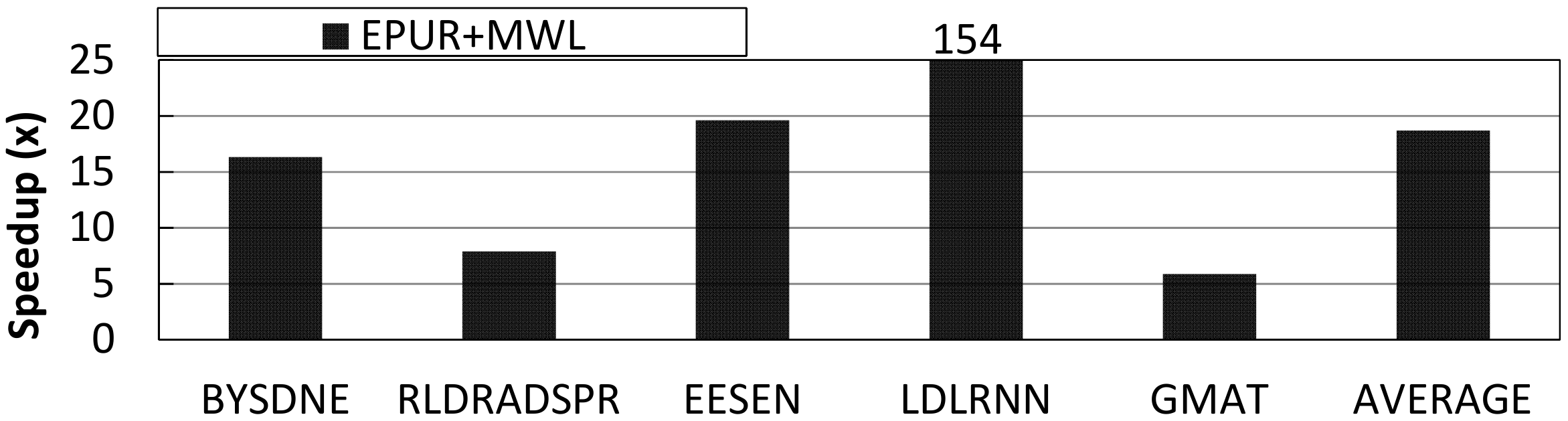}
	\caption{Speedups achieved by E-PUR over Tegra X1.}
	\label{f:speedup_tegra}
\end{figure}

Figure~\ref{f:speedup_tegra} shows the speedups for different LSTM
networks. On average, the speedup achieved by E-PUR
over Tegra X1 is 18.7x. E-PUR performance improvements come from hiding
memory latency (i.e, loading/storing is overlapped with computations), reducing off-chip memory accesses, and
featuring a custom pipeline tailored to
LSTM computation. Note that, for E-PUR, once the weights and input frames are loaded from the main system, 
there is not extra overhead from the main application. However, since the Tegra X1 is tailored to a broader range of 
applications, its performance is impacted by the overhead due to related tasks (e.g., GPU synchronization, CPU work, etc.).
Regarding E-PUR+MWL, there is not performance improvement against the baseline since the order in which MWL evaluates the
neurons does not change the final execution time. Note that in MWL, the number of operations to evaluate a given neuron 
is equal to the number of operations for the conventional order. However, because the evaluation of the recurrent connections for a given neuron is postponed until all forward connections are evaluated, the latency to evaluate a single neuron increases but the latency to produce the final output sequence does not change.
Finally, for speech recognition
applications, E-PUR achieves real-time performance by a
large margin, running 30x and 5x faster than real-time for
EESEN and RLDRADSPR respectively.

\begin{figure}[t!]
	\centering
	\includegraphics[width=3.375in]{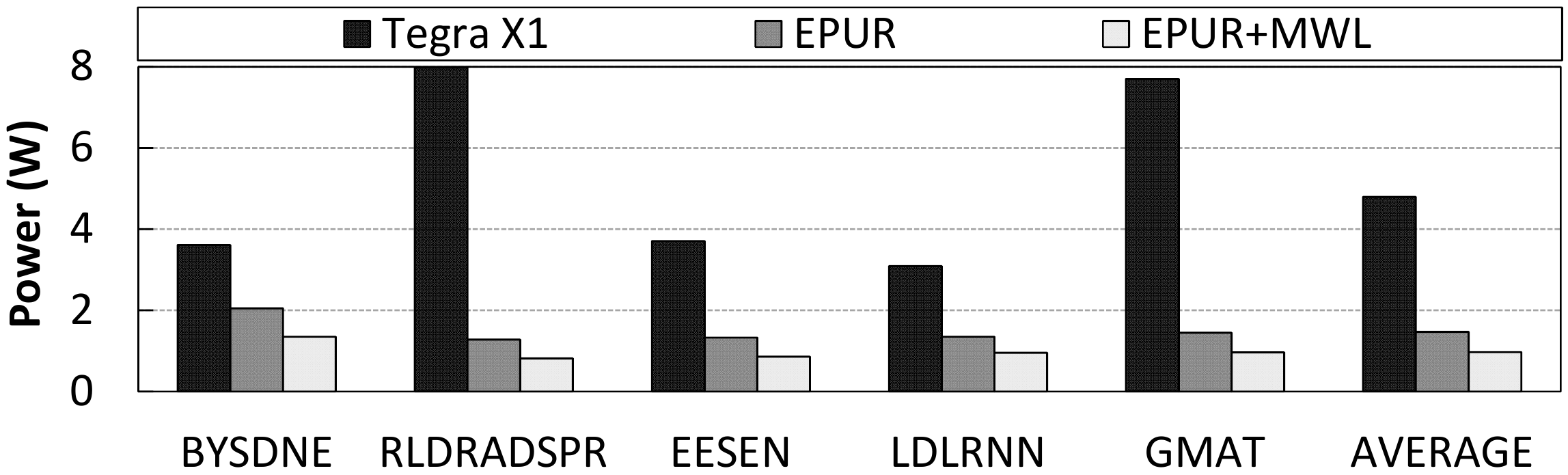}
	\caption{Power dissipation for E-PUR, E-PUR+MWL, and Tegra X1.}
	\label{f:power}
\end{figure}

On the other hand, power dissipation is shown in Figure~\ref{f:power}, which includes
the total power for Tegra X1 and the two configurations of
E-PUR. As it can be seen, E-PUR+MWL 
dissipates 5x lower power than Tegra X1 on average.

\begin{figure}[t!]
	\centering
	\includegraphics[width=3.375in]{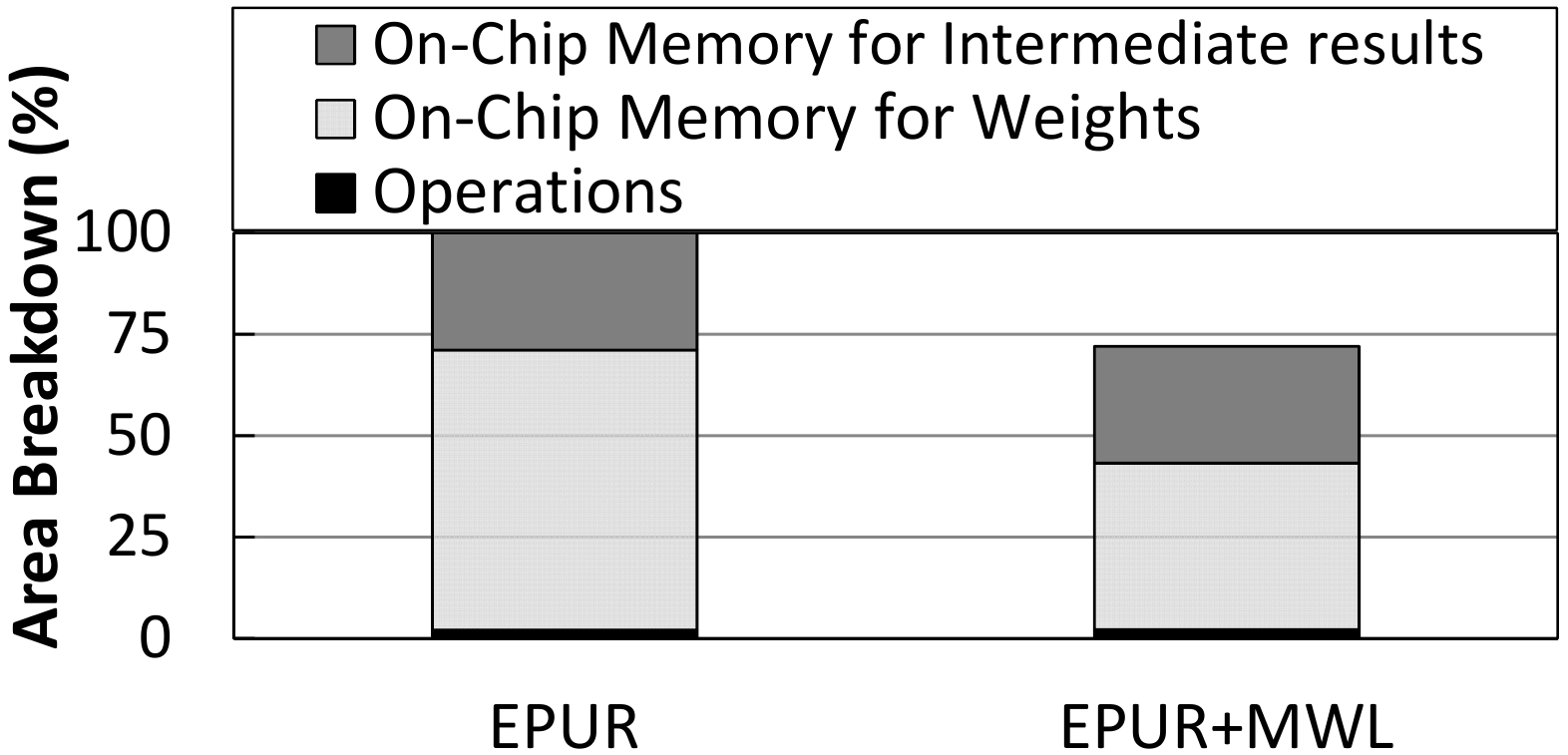}
	\caption{Normalized Area breakdown for E-PUR and E-PUR+MWL.}
	\label{f:total_area}
\end{figure}

Regarding area, E-PUR requires a total area of 64.6 $mm^2$,
whereas the total area of E-PUR+MWL is 46.3 $mm^2$. As
depicted in Figure~\ref{f:total_area}, the component with larger
contribution to the total area is the on-chip memory for the
synaptic weights, which is reduced by 50\% when MWL is applied.

\begin{figure}[t!]
	\centering
	\includegraphics[width=3.375in]{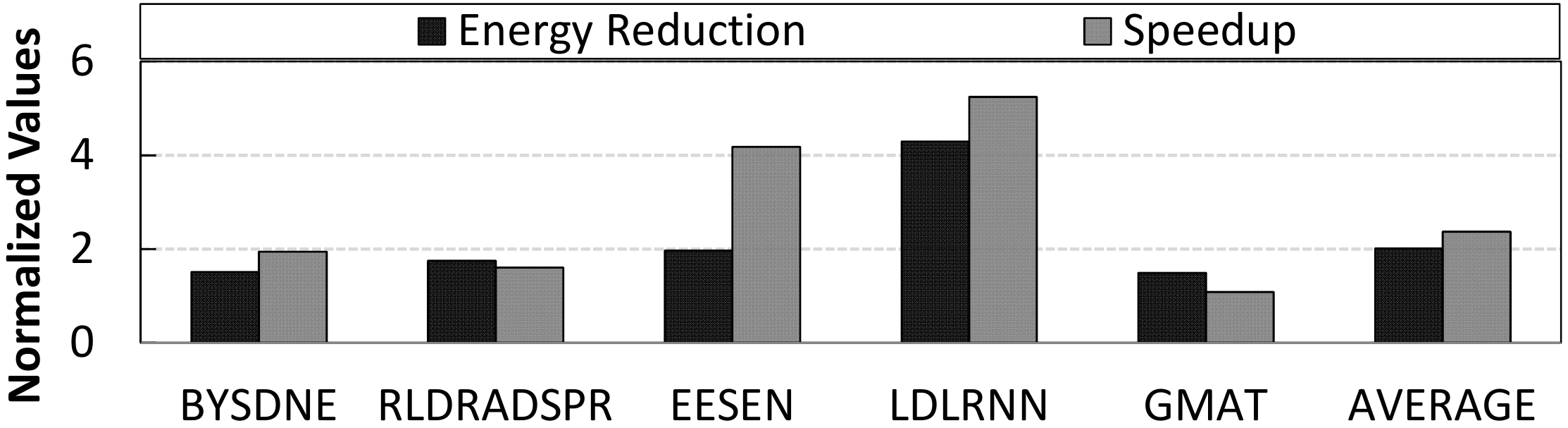}
	\caption{Speedup and energy reduction achieved by Tegra X1+MWL over Tegra X1.}
	\label{f:speedup_energy}
\end{figure}

Finally, Figure~\ref{f:speedup_energy} shows the speedup and energy reduction
of the Tegra X1+MWL, i.e. MWL implemented in
software, with respect to the baseline. On average, it
provides a 2x energy reduction and a 2.3x speedup.
EESEN and LDLRNN  exhibit large improvements
in performance and energy. These RNNs have smaller
number of neurons than the others (see Table~\ref{t:lstm_networks}) and, hence, the
synaptic weights can be stored in the on-chip storage of the
mobile GPU and reused for the entire layer evaluation, i.e. for
the whole input sequence. On the other hand, the
benefits are significantly smaller for BYSDNE,
RLDRADSPR and GMAT. These networks feature larger
number of neurons and, hence, the synaptic weights of one LSTM
cell cannot be stored on-chip in Tegra X1, increasing off-chip memory traffic by a
large extent. Note that the on-chip memories of Tegra X1
are fairly smaller than the ones included in E-PUR as
illustrated in Table~\ref{t:tegrax1} and Table~\ref{t:epur_params}.
This lack of on-chip storage
constrains the effectiveness of Tegra X1+MWL for RNNs
with large cell dimensionality.

%% file: related_work.tex
\section{Related Work}\label{s:related_work}

Improving the energy-efficiency of LSTM networks
has attracted the attention of the architectural community in
the last few years. Proposals for LSTM networks
acceleration have been presented in~\cite{han2017ese,guan2017fpga,li2015fpga}.
Although these accelerators achieve higher performance per watt than
CPUs and GPUs, they are not designed for low-power
mobile devices since their power dissipation ranges from
19 W to 41 W. However, E-PUR dissipates a
peak power of 970 mW, which is amenable for low-power
mobile devices.

Chang et al.~\cite{chang2015recurrent} present a low-power accelerator targeting the mobile segment. It implements a small LSTM network (2 layers, 128 neurons) and 
dissipates 1.9 W. In this work arithmetic operations are done using fixed-point Q8.8 data format, thus an accuracy loss of 7.1\% is aggregated. On the contrary, E-PUR uses floating point operations (either FP16 or FP32) and supports larger network models for a wide variety of application domains. Note that scaling up the aforementioned 
accelerator presented in~\cite{chang2015recurrent} to support larger LSTM networks would require a significant increase in local
storage capacity or in main memory traffic, and both alternatives would come at a high overhead in energy
consumption.

Another low-power LSTM accelerator is presented in~\cite{lee2016fpga},
this system consumes 9 W and supports larger models by
using aggressive weight quantization. External DRAM
traffic is completely avoided by storing the quantized
weights in a local on-chip memory of 2 Mbytes. However,
this quantization comes at the expense of non-negligible
accuracy loss. For speech recognition, Word Error Rate
increases from 13.5\%, using 32-bit floating point, to 15.1\%
and 20.2\% when using 6-bit and 4-bit quantization
respectively. Furthermore, larger and more accurate models
cannot be stored in its local memory even with the 4-bit
quantization. For example, EESEN requires more
than 5 Mbytes when using 4 bits per weight.
Our work is different since EPUR+MWL uses 8-bit quantization
to reduce the size of intermediate results with a
negligible impact on accuracy. 

The LSTM accelerator ESE~\cite{han2017ese} achieves high performance and energy-efficiency
by exploiting linear quantization and aggressive pruning. The main application for this work 
is speech recognition and the main target are high-end systems. On the contrary, E-PUR targets 
mobile devices and achieves high energy-efficiency by improving the
temporal locality of the memory accesses that fetch
synaptic weight. Moreover, E-PUR supports a large variety of applications.
We leave the use of pruned models in E-PUR as future work.

Regarding the work in~\cite{chen2016ESA}, E-PUR without MWL is similar to a weight stationary architecture applied to LSTMs since it loads all weights for given layer in on-chip memory, holding  them  until all associated computations are performed. However, MWL is different since it aims at further reducing the reuse distances. Unlike traditional weight stationary architectures, MWL splits synaptic weights in two types: forward and recurrent. Based on the observation that forward connections can be processed in any order, whereas recurrent connections impose sequential processing due to data dependencies. Therefore, MWL evaluates forward connections in the order that maximizes temporal locality, requiring extra small on-chip storage for this stage, whereas it processes all recurrent connections on a second stage as shown in Figure~\ref{f:mwl_reuse_distance}. MWL greatly reduces the energy consumption of the baseline accelerator.

Finally, cuDNN~\cite{cuDNN} has been recently extended to efficiently support RNN training. E-PUR design is significantly different in multiple ways. First, cuDNN focuses on RNN training with large batch sizes, whereas E-PUR focuses on RNN inference with batch size of one, i.e. one input sequence at a time. We measured cuDNN performance for RNN inference with batch size of one and found it is 1.5x faster than cuBLAS, whereas E-PUR achieves 18.7x speedup. cuDNN effectiveness is reduced due to the small batch size commonly used for RNN inference. Furthermore, cuDNN's optimizations to execute multiple layers in parallel cannot be applied to bidirectional LSTMs due to data dependencies.

%% file: conclusions.tex
\section{Conclusions}\label{s:conclusions}

In this paper, we present E-PUR, a processing unit for
RNNs that supports large LSTM networks while
dissipating low-power, motivated by the increasingly
important role of LSTM networks in applications such as speech
recognition, machine translation and video classification.
Unlike previous proposals that attempt to accommodate the
entire RNN on-chip, E-PUR only provides storage for one
LSTM layer, whose weights are fetched once from main
memory and reused for multiple recurrent executions. To
further improve the memory efficiency of E-PUR, we
introduce Maximizing Weight Locality (MWL), a novel
technique that improves the temporal locality of the
synaptic weights. The proposed design supports large
LSTM networks of hundreds of Megabytes, while
using small on-chip storage and low memory bandwidth.
Our results show that E-PUR reduces energy consumption
by 92x on average with respect to a modern mobile
GPU, while providing 18.7x speedup.